\documentclass[journal]{IEEEtran}

\usepackage{graphicx}
\usepackage{multirow}
\usepackage{multicol}
\usepackage{cite}
\usepackage{hyperref}
\usepackage{color}
\usepackage{amssymb}
\usepackage{url}
\usepackage{soul}
\usepackage{amsmath,amssymb,amsfonts}
\usepackage{algorithmic}
\usepackage{graphicx}
\usepackage{textcomp}
\usepackage{booktabs}
\begin{document}

\title{Equivariant Spherical CNNs for Accurate Fiber Orientation Distribution Estimation in Neonatal Diffusion MRI with Reduced Acquisition Time}

\author{Haykel Snoussi and Davood Karimi \\
\text{Department of Radiology, Boston Children's Hospital and Harvard Medical School, Boston, MA, USA}}

\maketitle

\begin{abstract}
Early and accurate assessment of brain microstructure using diffusion Magnetic Resonance Imaging (dMRI) is crucial for identifying neurodevelopmental disorders in neonates, but remains challenging due to low signal-to-noise ratio (SNR), motion artifacts, and ongoing myelination. In this study, we propose a rotationally equivariant Spherical Convolutional Neural Network (sCNN) framework tailored for neonatal dMRI. We predict the Fiber Orientation Distribution (FOD) from multi-shell dMRI signals acquired with a reduced set of gradient directions (30\% of the full protocol), enabling faster and more cost-effective acquisitions. We train and evaluate the performance of our sCNN using real data from 43 neonatal dMRI datasets provided by the Developing Human Connectome Project (dHCP). Our results demonstrate that the sCNN achieves significantly lower mean squared error (MSE) and higher angular correlation coefficient (ACC) compared to a Multi-Layer Perceptron (MLP) baseline, indicating improved accuracy in FOD estimation. Furthermore, tractography results based on the sCNN-predicted FODs show improved anatomical plausibility, coverage, and coherence compared to those from the MLP. These findings highlight that sCNNs, with their inherent rotational equivariance, offer a promising approach for accurate and clinically efficient dMRI analysis, paving the way for improved diagnostic capabilities and characterization of early brain development.
\end{abstract}

\begin{IEEEkeywords}
Diffusion MRI, spherical CNNs, neonatal neuroimaging, brain microstructure, geometric deep learning, tractography
\end{IEEEkeywords}

\section{Introduction}
\label{sec:introduction}
Medical diagnostics is undergoing a transformative shift, fueled by the rapid advancements in artificial intelligence (AI) and deep learning. These technologies are revolutionizing medical image analysis, offering the potential for unprecedented accuracy, speed, and automation in disease detection and diagnosis \cite{litjens2017survey}. Diffusion Magnetic Resonance Imaging (dMRI) is a non-invasive neuroimaging technique that provides unique insights into the microstructure of the brain and spinal cord tissue by measuring the diffusion of water molecules. By quantifying the directionality and magnitude of water diffusion, dMRI enables the mapping of white matter tracts and the characterization of microstructural changes associated with development \cite{snoussi2025haitch,karimi2024detailed}, aging \cite{luckey2024biological,snoussi2023diffusion}, and other neurodegenerative diseases \cite{snoussi2023effectiveness}. Accurate estimation of microstructural parameters from dMRI, particularly the Fiber Orientation Distribution (FOD), is crucial for early and precise diagnosis of neurodevelopmental disorders in neonates. Early detection can lead to timely interventions and improved outcomes. However, neonatal dMRI presents unique challenges, including small brain size, low SNR, motion artifacts, and the ongoing myelination process, which makes traditional analysis methods less reliable \cite{kebiri2024deep}.

The dMRI signal, denoted as \( E(\mathbf{q}) \), is acquired by applying diffusion-sensitizing gradients along various directions represented by the \(\mathbf{q}\)-vector. According to the Stejskal-Tanner equation, in the narrow pulse approximation (\(\delta << \Delta\), where \(\delta\) is the duration of the diffusion gradient and \(\Delta\) is the time between gradient pulses), \(E(\mathbf{q})\) is related to the Ensemble Average Propagator (EAP), \(P(\mathbf{r})\), through the Fourier transform \cite{stejskal1965spin}:

\begin{equation}
E(\mathbf{q}) = \int_{\mathbf{r} \in \mathbb{R}^{3}} P(\mathbf{r}) \exp(-2\pi i\mathbf{q} \cdot \mathbf{r}) \, d\mathbf{r}
\end{equation}

Here, \(P(\mathbf{r})\) represents the average probability of water molecules diffusing a distance \(\mathbf{r}\) over time \(\Delta\). A typical dMRI acquisition involves acquiring a reference image with no diffusion weighting (\(b = 0\) s/mm\(^2\)) and a series of diffusion-weighted images with varying \(\mathbf{q}\)-vectors.

Early identification of white matter abnormalities in neonates and accurately estimating microstructural parameters from dMRI are crucial for understanding brain architecture and identifying biomarkers for neurodevelopmental and neurological disorders. However, this task is challenging due to the inherent limitations of echo planar imaging (EPI) used in dMRI acquisition, such as geometric distortions and susceptibility to spin history effects, which distort the dMRI signals~\cite{hutter2018slice,snoussi2021evaluation,snoussi2019geometric,andersson2021diffusion}. This problem is further exacerbated in neonatal imaging due to the small brain size, low signal-to-noise ratio (SNR), and susceptibility to motion artifacts. Inconsistencies arising from these factors pose a major obstacle to accurately characterizing neonatal brain connectivity and hinder the overall reproducibility of dMRI studies~\cite{snoussi2025haitch,xiao2024reproducibility}. In addition, traditional approaches to extracting microstructural information from dMRI, such as multi-shell multi-tissue constrained spherical deconvolution (MSMT-CSD) \cite{jeurissen2014multi}, rely on fitting complex biophysical models to the dMRI signal, often requiring lengthy acquisitions and dense sampling schemes, which are less feasible in vulnerable neonatal populations. This reliance on extensive data acquisition presents significant challenges for the healthcare system, increasing scanning costs, limiting scanner throughput, and potentially delaying timely diagnosis and intervention for neonates with suspected neurological conditions.
 
Deep learning has emerged as a promising alternative for dMRI analysis, offering faster and potentially more robust parameter estimation~\cite{karimi2021deep,karimi2024detailed,kerkela2024spherical,kebiri2024deep}. Among these various methods, spherical convolutional neural networks (sCNNs) \cite{cohen2018spherical,esteves2018learning} have shown particular promise due to their inherent rotational equivariance. sCNNs are designed to be $\text{SO}(3)$-equivariant (i.e., rotating the input changes the output according to the same rotation) artificial neural networks that perform spherical convolutions with learnable filters. They enable rotationally equivariant processing of spherical data, making them well-suited for predicting microstructural parameters like the FOD from dMRI data.

\begin{figure}[!t]
\centerline{\includegraphics[width=\columnwidth]{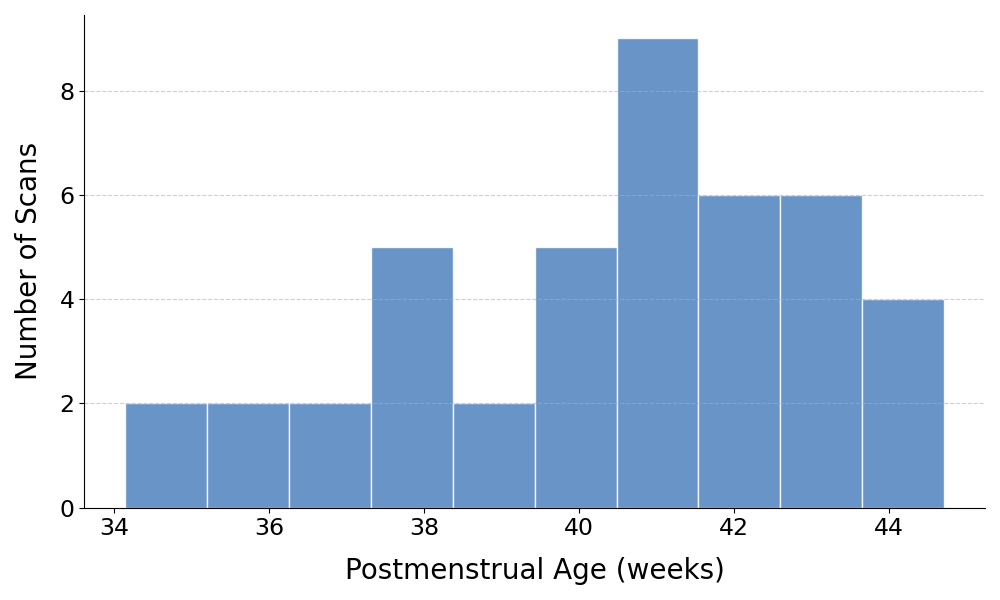}}
\caption{Distribution of the postmenstrual ages for the 43 neonatal dMRI datasets included in the study.}
\label{age_distribution}
\end{figure}

\begin{figure}[!t]
\centerline{\includegraphics[width=\columnwidth]{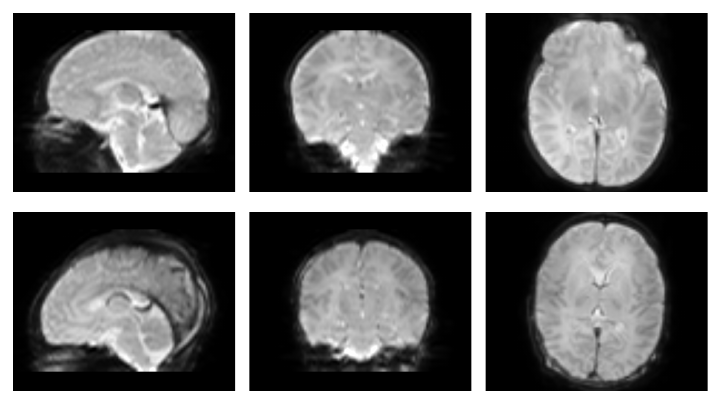}}
\caption{Sagittal, axial, and coronal views of representative examples data from neonatal dMRI in the dHCP dataset.}
\label{example_MRI}
\end{figure}

In this work, we develop and evaluate an sCNN framework for the challenging domain of neonatal dMRI, leveraging datasets from the Developing Human Connectome Project (dHCP) (see Figures \ref{age_distribution} and \ref{example_MRI}). We evaluate the performance of our framework using quantitative and qualitative metrics to demonstrate the downstream impact of accurate microstructural parameter estimation on connectomics analyses, a critical step towards clinical application. Our results demonstrate the improved robustness of the sCNN in estimating FODs in neonatal dMRI and therefore the tractography streamlines, compared to traditional approaches, paving the way for improved characterization of early brain development and, potentially, earlier and more accurate diagnosis of neurodevelopmental disorders. The complete implementation, including training scripts, model architectures, and evaluation tools, is publicly available at:
\url{https://github.com/H-Snoussi/sCNN-FOD-neonatal}. The main contributions of this work are:

\begin{enumerate}
\item Application and evaluation of the sCNN framework to a challenging and clinically relevant domain: neonatal dMRI.
\item Demonstrating that accurate FOD estimation is achievable using only 30\% of the full dHCP acquisition protocol, potentially enabling significantly faster, more cost-effective, and less burdensome neonatal dMRI scans.
\end{enumerate}

These contributions represent an important advance in investigating the broader applicability and robustness of the sCNN approach for microstructural parameter estimation in dMRI, specifically addressing the needs of neonatal neuroimaging and aligning with the goals of AI-driven medical image diagnostics.

\section{Materials and Methods}
\begin{figure*}[!t]
\centerline{\includegraphics[width=2.0\columnwidth]{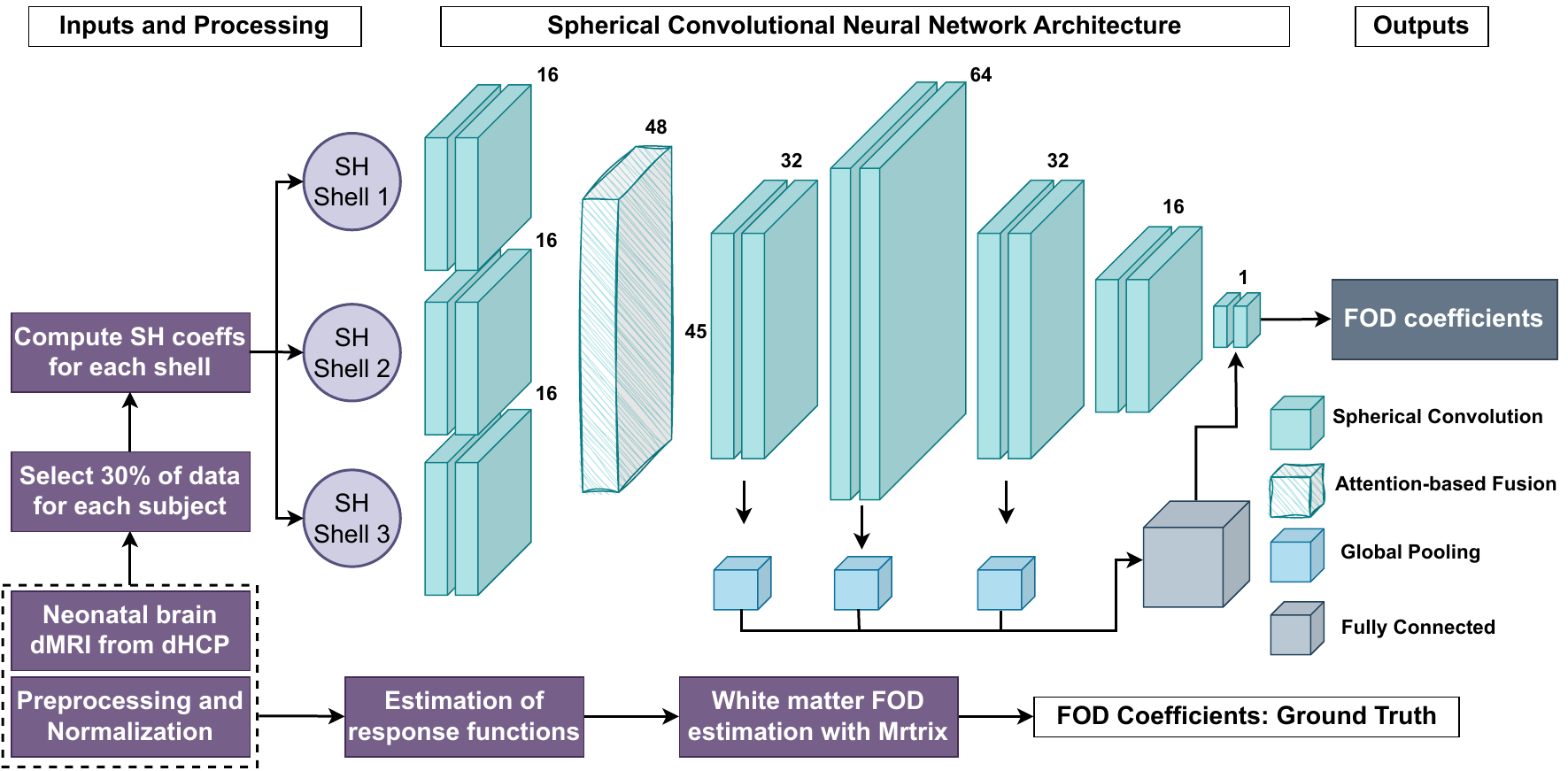}}
\caption{Flowchart illustrates the entire data processing and analysis pipeline, including the use of neonatal dMRI datasets, FOD estimation, data simulations, the sCNN architecture, and the outputs of the sCNN.}
\label{architecture}
\end{figure*}

The methodology employed in this study encompasses several key stages, from data representation and preprocessing to model development, training, and evaluation. A comprehensive overview of the entire process, including the processing of neonatal dMRI datasets, FOD estimation, sCNN architecture, and the network's outputs, is presented in Figure \ref{architecture}.

\subsection{Neonatal dMRI Data Acquisition and Preprocessing}
This study utilized 43 neonatal dMRI datasets from the Developing Human Connectome Project (dHCP). The postmenstrual age distribution of the included subjects is shown in Figure \ref{age_distribution}, and Figure \ref{example_MRI} provides two representative examples of the neonatal dMRI data.

The dHCP neonatal dMRI acquisition protocol was designed to optimize data acquisition for the unique properties of the developing brain. It employed a uniformly distributed set of gradient directions across three \(b\)-value shells \cite{edwards2022developing}. The protocol comprised 20 volumes at \(b=0\) s/mm\(^2\), 64 volumes at \(b=400\) s/mm\(^2\), 88 volumes at \(b=1000\) s/mm\(^2\), and 128 volumes at \(b=2600\) s/mm\(^2\). The temporal ordering of the acquired directions was strategically planned to maximize efficiency, mitigate the risks of infant motion artifacts, and adhere to gradient duty cycle constraints. Data were acquired with in-plane resolution of 1.5 \( \times \) 1.5 mm, and 3 mm slices with 1.5 mm overlap. Image reconstruction was performed using a dedicated algorithm \cite{hutter2018slice, cordero2018spin}.

The dataset underwent a comprehensive preprocessing pipeline, including denoising, brain masking, dynamic distortion correction, and slice-to-volume motion correction using a multi-shell spherical harmonics and radial decomposition (SHARD) representation \cite{christiaens2021scattered}. Simple intensity normalization was performed by setting negative values to zero and clipping high values at the 95th percentile.

\subsection{Ground Truth FOD Estimation}
FODs were estimated using the multi-shell multi-tissue constrained spherical deconvolution (MSMT-CSD) framework, as implemented in MRtrix3 \cite{tournier2019mrtrix3}. This approach decomposes the diffusion-weighted signal into contributions from white matter (WM), gray matter (GM), and cerebrospinal fluid (CSF) compartments \cite{jeurissen2014multi}. Response functions for each tissue type were estimated using the \textit{dhollander} algorithm \cite{dhollander2019improved}. The resulting WM FODs were extracted and used as ground truth for this study. These FODs were represented using spherical harmonics (SH) up to order \(lmax = 8\), resulting in 45 SH coefficients per voxel. The neonatal WM FOD datasets were randomly split into training (35 datasets), validation (4 datasets), and testing (4 datasets) sets.

\subsection{Generation of Reduced dMRI Training Data}
To facilitate faster and more cost-effective neonatal dMRI analysis, we generated training data using only the first 30\% of the full dHCP acquisition protocol's gradient directions. This reduced protocol consisted of 19 volumes at \(b=400\) s/mm\(^2\) (compared to 64 in the full protocol), 26 volumes at\(b=1000\) s/mm\(^2\) (compared to 88), and 38 volumes at \(b=2600\) s/mm\(^2\) (compared to 128). The \(b=0\) s/mm\(^2\) volumes are not considered in the computation of SH. For each \(b\)-value shell, spherical harmonic (SH) coefficients were extracted from the diffusion-weighted data up to \(lmax = 8\), resulting in 45 SH coefficients per shell.

\begin{table}[h]
\centering
\caption{Summary of dataset splits and number of diffusion directions. Ground truth FODs were estimated using the full acquisition (280 directions), while the sCNN was trained on reduced data (30\%, 83 directions).}
\begin{tabular}{lcccc}
\hline
\textbf{Split} & \textbf{Subjects} & \textbf{Voxels} & \multicolumn{2}{c}{\textbf{Diffusion MRI Directions}} \\
\cline{4-5}
              &                      &                    & \textbf{Ground Truth} & \textbf{Training} \\
\hline
Training      & 35                   & 3,703,986          & 280                   & 83                \\
Validation    & 4                    & 358,015            & 280                   & 83                \\
Testing       & 4                    & 483,950            & 280                   & 83                \\
\hline
\end{tabular}
\label{table:data_summary}
\end{table}

\subsection{sCNN Model for FOD Estimation}
The core of this study is a Spherical Convolutional Neural Network (sCNN) designed to estimate WM FOD from a reduced set of dMRI measurements. The sCNN architecture is optimized for spherical signals, leveraging spherical convolutions to exploit the rotational properties of diffusion signals. This approach ensures a more structured and efficient learning process, maintaining consistency across different orientations.

\subsubsection{sCNN Architecture and Shell Attention Mechanism}
The proposed sCNN model is built upon a hierarchical, shell-specific feature extraction strategy, incorporating attention mechanisms to enhance feature fusion across different diffusion shells. The architecture is illustrated in Figure \ref{architecture}.

Shell-specific convolutions are applied independently to the input diffusion-weighted data at different shells using three spherical convolutional layers. Each layer extracts relevant features from its corresponding shell before passing them to the next stage. To improve feature integration across shells, a shell attention module is employed, assigning dynamic weights to different shells to enhance the learning of critical structures by prioritizing the most informative features.

Shell-specific convolutions are applied independently to the input diffusion-weighted data at different shells using three spherical convolutional layers. Each layer extracts relevant features from its corresponding shell before passing them to the next stage. To improve feature integration across shells, a shell attention module is employed, assigning dynamic weights to different shells to enhance the learning of critical structures by prioritizing the most informative features. Specifically, for each shell-specific feature map \( \mathbf{X}_i \in \mathbb{R}^{B \times 16 \times C} \) (where \( B \) is the batch size and \( C \) is the number of spherical harmonic coefficients), global average pooling is applied across the SH dimension to form a 48-dimensional feature vector \( \mathbf{z} \in \mathbb{R}^{B \times 48} \) by concatenating  $\mathbf{z} = \left[ \text{mean}(\mathbf{X}_1), \text{mean}(\mathbf{X}_2), \text{mean}(\mathbf{X}_3) \right]$. The resulting feature vector \( \mathbf{z} \in \mathbb{R}^{48} \) is passed through a two-layer feedforward network to generate shell attention logits \( \mathbf{l} \in \mathbb{R}^3 \):
\begin{equation}
\mathbf{l} = \mathbf{W}_2 \cdot \sigma\left( \mathbf{W}_1 \mathbf{z} + \mathbf{b}_1 \right) + \mathbf{b}_2,
\end{equation}
where \( \mathbf{W}_1 \in \mathbb{R}^{24 \times 48} \), \( \mathbf{W}_2 \in \mathbb{R}^{3 \times 24} \), and \( \sigma(\cdot) \) is a Leaky Rectified Linear Unit (Leaky ReLU) non-linearity with negative slope 0.1. The attention weights \( \mathbf{a} \in \mathbb{R}^3 \) are then computed using the softmax function:
\begin{equation}
a_i = \frac{\exp(l_i)}{\sum_{j=1}^3 \exp(l_j)} \quad \text{for } i = 1, 2, 3,
\end{equation}
ensuring that \( \sum_i a_i = 1 \) and \( a_i \geq 0 \). These weights are broadcast and applied multiplicatively to each shell-specific feature map before concatenation. This mechanism enables the model to assign higher importance to more informative shells on a per-sample basis, rather than treating all shells equally.

Following attention-guided fusion, the network applies a series of spherical convolutional layers in an encoder-decoder configuration with increasing feature channels: 16, 32, and 64. Leaky ReLU activation functions are applied after each layer to introduce non-linearity. The decoder progressively refines the feature representations using a symmetric series of spherical convolutions, which enhances feature retention and improves reconstruction quality. Finally, the processed feature maps are passed through fully connected layers with batch normalization and ReLU activations to enhance learning efficiency. The output layer produces 45 SH coefficients representing the estimated WM FODs.

\subsubsection{Rotationally Equivariant Spherical Convolution Layers}
The foundational operation in our sCNN architecture is the spherical convolution, which is specifically designed to process functions defined on the sphere—such as the diffusion MRI (dMRI) signal—while preserving rotational structure. In diffusion imaging, signals are naturally represented using spherical harmonics (SH), a basis for functions on the unit sphere. SH coefficients capture both the magnitude and directionality of signal variation, making them particularly well-suited for modeling fiber orientation distributions.

Mathematically, a spherical convolution between a function \( f \) and a filter \( h \) is defined as:
\begin{equation}
(f * h)(\mathbf{x}) = \int_{\text{SO}(3)} \text{d}\mathbf{R} \, f(\mathbf{R} \hat{\mathbf{e}}_3) \, h(\mathbf{R}^{-1} \mathbf{x}),
\end{equation}
where \( \mathbf{x} \) is a point on the sphere, \( \hat{\mathbf{e}}_3 \) is the north pole unit vector, and \( \mathbf{R} \in \text{SO}(3) \) denotes a rotation. This operation is equivariant to 3D rotations, meaning:
\[
\text{If } f'(\mathbf{x}) = f(\mathbf{R}^{-1} \mathbf{x}), \text{ then } (f' * h)(\mathbf{x}) = (f * h)(\mathbf{R}^{-1} \mathbf{x}),
\]
so rotating the input results in a rotated output. This is a critical property for diffusion MRI analysis, where fiber orientations can vary arbitrarily in space.

In our implementation, the spherical convolution is performed directly in the SH domain. Each degree \( l \) is associated with a learnable scalar weight that is shared across all \( m \)-orders within that degree. This ensures that the operation is SO(3)-equivariant, as rotations in SH space only mix coefficients within the same degree. These weights are stored in a tensor of shape \( [C_{\text{out}}, C_{\text{in}}, L] \), where \( L \) is the number of SH degrees (restricted to even \( l \) for antipodal symmetry, as is standard in diffusion MRI). A degree expansion mask is used to broadcast these scalar weights to all orders \( m \), and the convolution is applied using an efficient Einstein summation.

To introduce non-linearity while preserving spherical structure, SH coefficients are transformed to the spatial domain using the Inverse Spherical Fourier Transform (ISFT), followed by a Leaky ReLU activation and then mapped back to the SH domain using the forward the Spherical Fourier Transform (SFT). While this spatial-domain nonlinearity breaks strict SO(3) equivariance, it preserves approximate rotation-awareness and maintains compatibility with the SH-based structure of the data.

When the number of input and output channels match, a residual connection is applied, which is inherently equivariant since addition is commutative with rotation. Only even SH degrees are used (e.g., \( l = 0, 2, 4, \ldots \)), reflecting the antipodal symmetry of diffusion signals and reducing unnecessary parameters.

In summary, our spherical convolution layers apply band-limited, degree-wise learnable weights in the SH domain, preserving SO(3)-equivariance. Approximate equivariant nonlinearities are applied via ISFT/SFT transformations, ensuring the network remains lightweight and robust to arbitrary signal orientations. This design enables biologically and physically informed feature learning, critical for accurate and generalizable fiber orientation estimation in dMRI.

\subsubsection{Spatial Domain Loss Function for FOD Reconstruction}
Standard Mean Squared Error (MSE) loss, when applied directly to Spherical Harmonic (SH) coefficients, is suboptimal for FOD reconstruction. This is because SH coefficients do not contribute equally to the reconstructed FOD. Lower-order coefficients primarily govern the overall magnitude or isotropic component, while higher-order coefficients capture finer angular details. Using a basic MSE loss treats all coefficients equally, potentially penalizing errors in higher-order coefficients less than errors in lower-order ones, even though the latter can have a more significant impact on the overall FOD shape. Therefore, a more nuanced approach is required. We propose a modified MSE loss calculated in the spatial domain, rather than the SH domain, to address this issue.

Specifically, given predicted SH coefficients $\mathbf{p} \in \mathbb{R}^{B \times 45}$ and target SH coefficients $\mathbf{t} \in \mathbb{R}^{B \times 45}$ for a batch of size $B$ the loss function first reconstructs the FOD signals in the spatial domain using the ISFT:

\begin{equation}
    \mathbf{p}_{\text{FOD}} = \mathbf{U} \mathbf{p}, \quad \mathbf{t}_{\text{FOD}} = \mathbf{U} \mathbf{t}
\end{equation}

where $\mathbf{U} \in \mathbb{R}^{N \times 45}$ is the ISFT matrix mapping SH coefficients back to the spatial domain, $N$ is the number of spatial points used to represent the FOD. The loss is then computed as the mean squared difference between the predicted and target FOD signals:

\begin{equation}
    \mathcal{L}_{\text{MSE}} = \frac{1}{N B} \sum_{i=1}^{B} \sum_{j=1}^{N} \left( \mathbf{p}_{\text{FOD},ij} - \mathbf{t}_{\text{FOD},ij} \right)^2
\end{equation}

By computing the loss in the spatial domain rather than directly in the SH coefficient space, this approach ensures that model predictions are optimized for their impact on the reconstructed diffusion signal rather than just the coefficient magnitudes. This strategy improves the model's ability to generate accurate fiber orientation estimates.

\subsubsection{Training Procedure}
The training procedure of the sCNN model was designed to optimize convergence while preventing overfitting. The model was trained using the AdamW optimizer with an initial learning rate of \( 10 e -4\) and a weight decay of \( 10 e -4\). The learning rate was adjusted using a step-based scheduler with a decay factor of 0.5 every 17 epochs. To ensure stable training, gradient clipping was applied with a maximum norm of 10.0.

Training data consisted of diffusion-weighted images sampled from a reduced set of gradient directions, from which spherical harmonic (SH) coefficients were extracted up to \(lmax = 8\), resulting in 45 SH coefficients per voxel. The model was trained for 80 epochs for one hour. The MSE loss function was used, computed after transforming the SH coefficients into the spatial domain using ISFT.

\subsection{Comparison with Multi-Layer Perceptron (MLP):}
We compared the performance of the sCNN with a common deep learning network, Multi-Layer Perceptron (MLP)~\cite{goodfellow2016deep}. We trained an MLP with four fully connected layers (256 nodes each) followed by batch normalization and ReLU activations. The MLP took the normalized dMRI signals as input and output the spherical harmonic coefficients of the FOD. The MLP was trained using MSE as loss function and optimizer as the sCNN but required five times more training batches to ensure convergence due to its higher parameter count. Despite its simplicity, the MLP provided a baseline for assessing the effectiveness of spherical convolutions in capturing rotationally invariant features.

\subsection{Evaluation Metrics}
To comprehensively evaluate the performance of the proposed sCNN model comparing to the ground truth and the baseline method, we employed a set of quantitative and qualitative metrics. These metrics were designed to assess both the accuracy of the estimated FODs and their downstream impact on WM tractography. The quantitative metrics include MSE, Angular Correlation Coefficient (ACC), and Structural Similarity Index Measure (SSIM), which evaluate how closely the predicted FODs match the ground truth in both coefficient and angular space. Additionally, we conducted tractography-based assessments to evaluate the practical implications of FOD quality on the reconstruction of WM pathways.

\subsubsection{Mean Squared Error (MSE)}
The MSE was used as the primary metric to quantify the discrepancy between the predicted and reference FODs. For each voxel, the MSE was computed directly in the SH domain as the mean squared difference between the predicted and ground truth SH coefficients:
\[
\text{MSE} = \frac{1}{N} \sum_{i=1}^{N} \| \hat{S}_i - S_i \|^2,
\]
where \( \hat{S}_i \) and \( S_i \) are the predicted and ground truth SH coefficient vectors for voxel \( i \), and \( N \) is the number of voxels.

\subsubsection{Angular Correlation Coefficient (ACC)}
The Angular Correlation Coefficient (ACC) measures the similarity in orientation between predicted and ground truth FODs~\cite{anderson2005measurement}. For each voxel, FODs are reconstructed by projecting SH coefficients onto a dense spherical grid. ACC is then calculated as the cosine similarity between the reconstructed FODs:
\[
\text{ACC} = \frac{\langle \hat{FOD}, FOD \rangle}{\| \hat{FOD} \| \cdot \| FOD \|},
\]
where \( \hat{FOD} \) and \( FOD \) represent the predicted and ground truth FOD amplitudes over the sphere.

\subsubsection{Structural Similarity Index Measure (SSIM)} 
The Structural Similarity Index (SSIM) is a perceptual metric that quantifies the similarity between two images, considering luminance, contrast, and structural information. In the context of FOD evaluation, we compute SSIM specifically on the zeroth-order spherical harmonic (SH0) coefficient, which reflects the isotropic component of the FOD signal and is sensitive to the overall diffusion strength. This metric provides a complementary assessment to MSE, focusing on the perceptual and structural similarity of low-frequency diffusion content between predicted and reference FODs.

\subsubsection{Tractography-Based Evaluation}
To assess the downstream utility of the predicted FODs, we performed probabilistic tractography using the iFOD2 algorithm \cite{tournier2010improved} implemented in MRtrix3 \cite{tournier2019mrtrix3}. Streamlines were seeded uniformly throughout a white matter mask and generated using the predicted FODs. The resulting tractograms were visually inspected for anatomical plausibility, coherence, and coverage of major white matter bundles. This qualitative evaluation helps determine whether differences in FOD estimation affect tract reconstruction.

\subsection{Implementation Details and Code Availability} 
The sCNN and MLP models were implemented using PyTorch \cite{paszke2019pytorch} and trained on an NVIDIA RTX A6000 GPU with 48 GB of memory. Training the sCNN model required approximately 1.1 hours, while the MLP model required approximately 6 hours. The source code, trained models, and scripts for reproducing the results are publicly available at \url{https://github.com/H-Snoussi/sCNN-FOD-neonatal}.

\section{Experiments and Results}
\subsection{Quantitative Evaluation of FOD Estimation Accuracy}
Table \ref{tab:eval_metrics_extended} presents the quantitative results for FOD estimation, comparing the sCNN and MLP models against the ground truth (MSMT-CSD) on the test set. The rotationally equivariant sCNN significantly outperformed the MLP in all metrics. The sCNN achieved an 84\% reduction in the MSE compared to the MLP (0.0017 vs 0.0108), indicating superior reconstruction of SH coefficients. The sCNN also demonstrated a higher ACC (18.67° vs. 8.68°), signifying better capacity to capture angular information in the FOD, particularly in regions with complex fiber architecture. The higher ACC variability for sCNN (±32.1°) versus MLP (±9.5°) reflects its sensitivity to true anatomical complexity rather than noise - while MLP collapses to mean orientation estimates, the sCNN preserves genuine but variable crossing fibers (Fig.~\ref{FOD_zoom_roi}). Furthermore, the SSIM computed on the SH0 term was substantially higher for the sCNN (0.3453 ± 0.3403) than for the MLP (0.0588 ± 0.0285), indicating better preservation of the isotropic diffusion component, crucial for distinguishing developing white matter from unmyelinated regions.

\begin{table}[ht]
\caption{Evaluation metrics (MSE, ACC, SSIM) averaged over test subjects (mean ± std).}
\label{tab:eval_metrics_extended}
\centering
\begin{tabular}{lccc}
\toprule
\textbf{Method} & \textbf{MSE} & \textbf{ACC} & \textbf{SSIM (SH0)} \\
\midrule
MLP   & 0.0108 ± 0.0039 & 8.6844 ± 9.4905  & 0.0588 ± 0.0285 \\
sCNN  & 0.0017 ± 0.0010 & 18.6708 ± 32.1065 & 0.3453 ± 0.3403 \\
\bottomrule
\end{tabular}
\end{table}

\begin{figure*}[!t]
\centerline{\includegraphics[width=2.15\columnwidth]{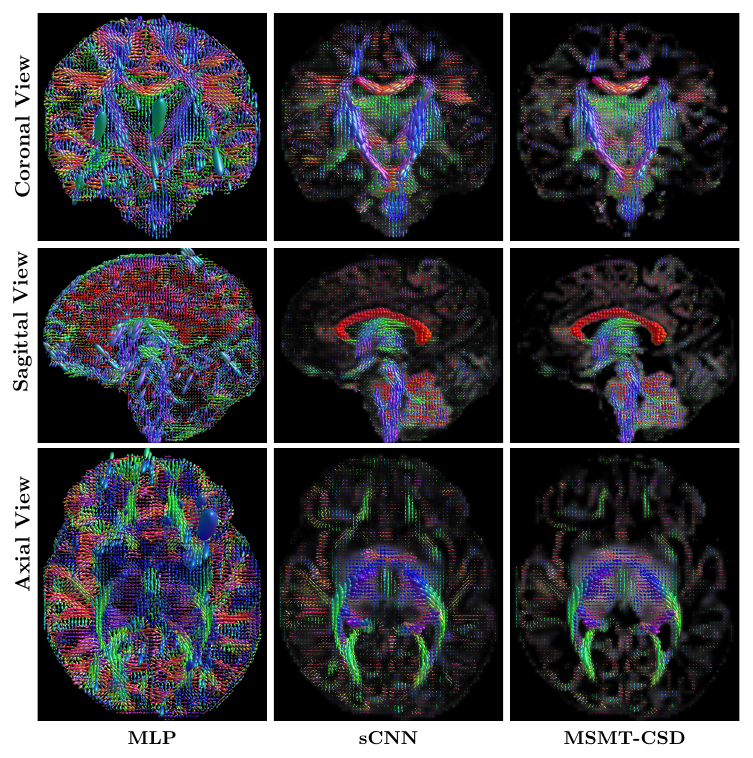}}
\caption{Representative FODs from a test subject. (left column) FODs estimated by the MLP using the full dHCP dataset. (middle column) FODs estimated by the sCNN using 30\% of the diffusion directions. (right column) Ground truth FODs estimated using MSMT-CSD with the full dHCP dataset. The sCNN produces FODs that are visually much more similar to the ground truth than the MLP.}
\label{FOD_full}
\end{figure*}

\begin{figure*}[!t]
\centerline{\includegraphics[width=2.15\columnwidth]{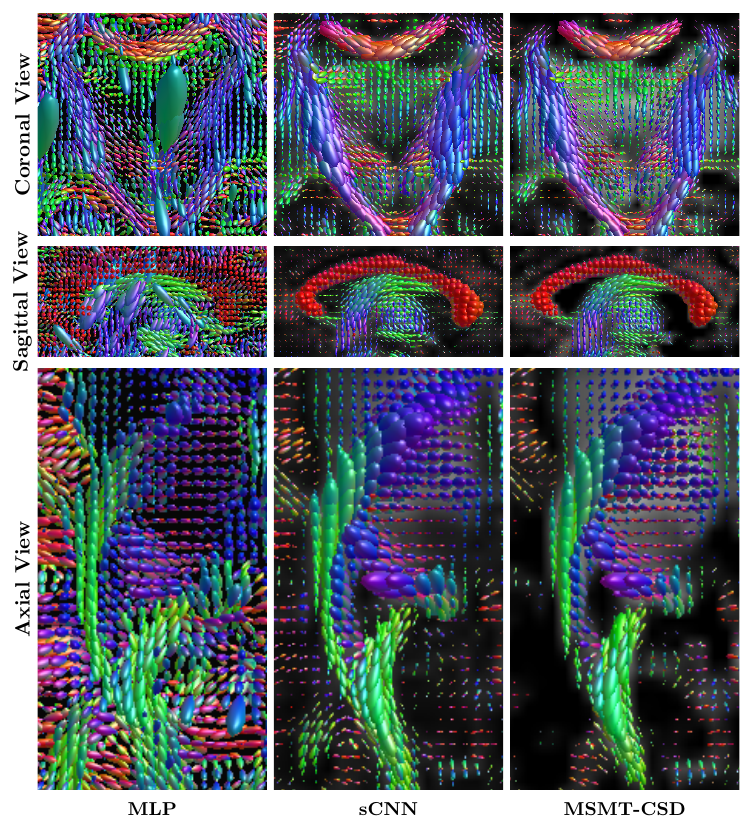}}
\caption{Zoomed-in views of regions of interest (ROIs) with complex fiber configurations, highlighting differences between FODs predicted by MLP, sCNN, and MSMT-CSD (ground truth). The sCNN preserves anatomical structure and closely resembles the ground truth, whereas the MLP exhibits increased noise and reduced structural clarity. These ROIs correspond to those shown in Figure~\ref{FOD_full}.}
\label{FOD_zoom_roi}
\end{figure*}

\subsection{Qualitative FOD Visualization}
Figure \ref{FOD_full} and Figure \ref{FOD_zoom_roi} present visual comparisons of the FODs estimated by the sCNN, MLP, and MSMT-CSD (ground truth) for an example of a test subject. Visually, the sCNN-predicted FODs closely resemble those generated by MSMT-CSD, demonstrating clear and anatomically consistent fiber peaks with reduced noise and spurious orientations. In contrast, the MLP FODs appear noisy and less defined, with many spurious peaks and a lack of clear directional coherence in regions with complex fiber crossings. The sCNN FODs show sharper peaks and better delineate fiber orientations. Figure \ref{FOD_zoom_roi} provides a zoomed-in view of specific regions of interest (ROI) to further show the superior performance of the rotationally equivariant sCNNs in preserving structural fidelity and resolving complex fiber architectures. These visualizations corroborate the quantitative improvements and suggest that the sCNN better preserves complex microstructural features critical for reliable downstream analyses.

\subsection{Tractography Analysis}
\begin{figure*}[!t]
\centerline{\includegraphics[width=2.15\columnwidth]{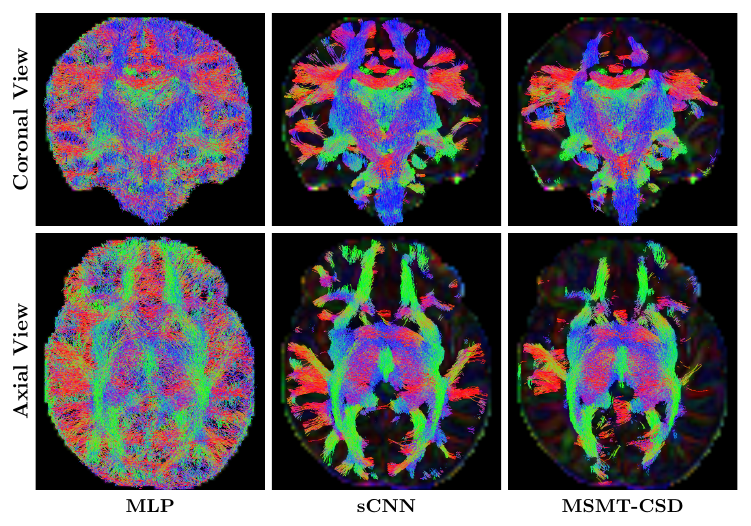}}
\caption{Representative tractography results. (left) Tractogram generated using MLP-predicted FODs. (middle) Tractogram generated using sCNN-predicted FODs. (right) Tractogram generated using ground truth FODs (MSMT-CSD).}
\label{tracto_fig}
\end{figure*}

Tractography, while a powerful tool for visualizing white matter pathways, is inherently sensitive to the quality of the underlying FOD estimates. Figure~\ref{tracto_fig} shows tractography results generated from the FODs produced by each method. Remarkably, the sCNN-based tractograms demonstrate superior anatomical plausibility, coherence, and completeness compared to both the MLP-based tractograms and, notably, the tractograms generated from the MSMT-CSD ground truth FODs. The sCNN successfully reconstructs major white matter pathways, such as the corpus callosum, and the corticospinal tract, with greater fidelity and fewer spurious streamlines than both other methods. The MLP tractogram exhibits significant noise and fails to accurately represent these key pathways. This suggests that the sCNN not only learns to estimate FODs but also effectively denoises and improves the underlying representation of white matter pathways, leading to more robust and reliable tractography.

\section{Discussion}
This study demonstrates the significant potential of rotationally equivariant sCNN for accurate and efficient FOD estimation in neonatal dMRI, using a substantially reduced acquisition protocol. Our results show that the proposed sCNN model not only significantly outperforms a standard MLP in terms of MSE, ACC, and SSIM, indicating superior FOD estimation accuracy, but also produces results that surpass those obtained using the conventionally accepted MSMT-CSD ground truth. These findings have important implications for the analysis of neonatal dMRI data and the potential for earlier, more accurate, and more efficient diagnosis of neurodevelopmental disorders.

\subsection{Model Design and Performance Drivers}
The superior performance of the sCNN is attributable to several key factors inherent to its design. First, the sCNN's core property of rotational equivariance ensures that it learns features that are intrinsically invariant to the orientation of the head within the scanner. This is a fundamental requirement for dMRI analysis, as the diffusion signal's orientation directly reflects the underlying fiber orientation. The MLP, lacking this built-in equivariance, must learn rotational invariance from the data, which is a more challenging task that typically requires larger datasets and more complex model architectures. Second, the shell attention mechanism allows the sCNN to dynamically weight the contributions of different \(b\)-value shells, which may vary depending on the degree of myelination. Third, the use of spherical convolutions allows the sCNN to operate directly on the SH representation of the diffusion signal. This avoids the need for interpolation or resampling, which can introduce artifacts and degrade the accuracy of FOD estimation. Fourth, the spatial-domain loss function, computed after transforming the SH coefficients to the spatial domain, emphasizes perceptually all SH orders without neglecting the finer angular details captured by lower-order coefficients. This ensures that the model optimizes for the shape of the FOD, not just the SH coefficient values. Finally, and critically, the sCNN's convolutional nature implicitly introduces spatial regularization. Unlike MSMT-CSD and the MLP, which treat each voxel independently, the sCNN considers a neighborhood of voxels during each convolution. This effectively smooths the estimated FODs, reducing the impact of noise and promoting anatomically plausible fiber orientations, while still preserving sharp details at fiber crossings and boundaries due to the learnable nature of the convolutional filters.

\subsection{Tractography and Diagnostic Quality}
The most striking finding of this study is the superior quality of the sCNN-based tractography compared to both the MLP-based tractography and the tractography generated from the MSMT-CSD ground truth FODs. The sCNN tractograms exhibit greater anatomical plausibility, improved delineation of major WM pathways, and, crucially, fewer spurious streamlines and less fragmentation than even the MSMT-CSD results. This suggests that the sCNN is not simply learning to mimic the MSMT-CSD output; it is, in effect, learning a better representation of the underlying WM architecture. This likely stems from the sCNN's ability to effectively denoise the diffusion signal and learn robust features from the reduced dataset, mitigating the limitations of conventional model-based methods like MSMT-CSD, which can be susceptible to noise and model misspecification.

\subsection{Clinical Impact and Translational Relevance}
The fact that accurate FOD estimation and superior tractography can be achieved using only 30\% of the full dHCP acquisition protocol has substantial practical implications. Reducing scan time is crucial in neonatal imaging, as it improves patient comfort, minimizes the risk of motion artifacts, and increases scanner availability, making dMRI more accessible for routine clinical use. This finding underscores the sCNN's ability to extract more information from a limited amount of data, a critical advantage in challenging imaging scenarios. Our reduced acquisition protocol holds strong potential for enabling unsedated scanning during natural sleep cycles, critical for monitoring preterm infants at risk for cerebral palsy. This could triple scanner throughput in the Neonatal Intensive Care Units (NICUs) while reducing parental anxiety from prolonged separations.
\subsection{Limitations and Future Work}
Despite these promising results, this study has some limitations. While the sample size of 43 subjects is larger than many previous studies in dMRI, future work should validate these findings on larger, more diverse datasets, including subjects with different clinical conditions. Second, while MSMT-CSD is a widely accepted method for FOD estimation, it is not a perfect gold standard. Future research should explore comparisons with other ground truth methods, including histological validation or advanced biophysical models, although obtaining such ground truth data in neonates is exceptionally challenging. Third, the reduced acquisition protocol used in this study (30\% of directions) was chosen empirically. Future work should investigate the optimal acquisition protocol for sCNN-based FOD estimation, potentially using active learning strategies to identify the most informative diffusion directions.
\subsection{Broader Applicability to Medical Diagnostics}
Beyond the immediate application to neonatal dMRI, our findings suggest that sCNNs have broader potential for improving dMRI analysis in other populations and applications. The challenges of motion artifacts and scan time constraints are even more pronounced in fetal and pediatric dMRI, making the sCNN approach potentially even more valuable in these contexts. The public release of our training pipeline, including the trained models and data processing scripts, facilitates the rapid translation of this technology to other vulnerable populations and encourages further research in this area. Moreover, this work aligns closely with the goals of AI-driven medical diagnostics by demonstrating how deep learning architectures can enhance image-derived biomarkers, improve diagnostic accuracy, and enable earlier interventions in clinical neuroimaging workflows.

\section{Conclusion}
This study contributes to the growing body of research on deep learning for medical image diagnostics by demonstrating the feasibility and potential of sCNNs for accurate and efficient FOD estimation in neonatal dMRI. The proposed sCNN model outperforms a standard MLP in terms of both quantitative metrics and tractography results, highlighting the benefits of rotational equivariance and shell-specific processing. The ability to achieve accurate FOD estimation with a reduced acquisition protocol has significant implications for clinical practice, potentially leading to faster, more cost-effective, and less burdensome neonatal dMRI scans. This research paves the way for improved characterization of early brain development and earlier, more accurate diagnosis of neurodevelopmental disorders, contributing to improved clinical outcomes for vulnerable neonatal populations. 

\section*{Acknowledgment}
This research was supported in part by the National Institutes of Health (NIH) under award number R01HD113199.

% \section*{References}
\bibliographystyle{IEEEtranS}
\bibliography{references}

\end{document}